\documentclass[11pt]{article}

\usepackage[T1]{fontenc}
\usepackage[utf8]{inputenc}

\usepackage[a4paper,margin=1in]{geometry}
\usepackage{microtype}

\usepackage{amsmath,amssymb,amsthm}

\newtheorem{proposition}{Proposition}

\usepackage{graphicx}
\usepackage{booktabs}
\usepackage{tabularx}
\usepackage{multirow}
\usepackage{adjustbox}
\usepackage{subcaption}

\usepackage{enumitem}
\usepackage{xspace}
\usepackage{comment}
\usepackage{xcolor}

\usepackage{algorithm}
\usepackage{algorithmicx}
\usepackage{algpseudocode}

\usepackage{tikz}
\usetikzlibrary{positioning,arrows.meta,shapes,calc}
\usepackage{pgfplots}
\pgfplotsset{compat=1.17}

\usepackage{placeins}
\usepackage{float}

\usepackage[colorlinks=true,allcolors=blue]{hyperref}

\usepackage{authblk}

\newcommand{\keywords}[1]{\par\noindent\textbf{Keywords:} #1\par}

\title{Learning to Trust Experience: A Monitor--Trust--Regulator Framework for Learning under Unobservable Feedback Reliability}

\author[1,2]{Zhipeng Zhang}
\author[3]{Zhenjie Yao}
\author[1]{Kai Li}
\author[1]{Lei Yang}
\affil[1]{China Mobile Research Institute, Beijing 100053, China}
\affil[2]{China Mobile GBA (Greater Bay Area) Innovation Institute, Guangzhou 510656, China}
\affil[3]{Institute of Microelectronics, Chinese Academy of Sciences, Beijing 100029, China}

\date{} % remove date

\begin{document}
\maketitle

\begin{center}
\textbf{Corresponding author:} Zhipeng Zhang (\texttt{zhangzhipeng@chinamobile.com})
\end{center}

%%%%%%%%%%%%%%%%%%%%%%%%%%%%%%%%%%%%%%%%%%%%%%%%%%%%
% Abstract
%%%%%%%%%%%%%%%%%%%%%%%%%%%%%%%%%%%%%%%%%%%%%%%%%%%%
\begin{abstract}
Learning under unobservable feedback reliability poses a distinct challenge beyond optimization robustness: a system must decide \textit{whether} to learn from an experience, not only \textit{how} to learn stably. We study this setting as \textbf{Epistemic Identifiability under Unobservable Reliability (EIUR)}, where each experience has a latent credibility, reliable and unreliable feedback can be locally indistinguishable, and data are generated in a closed loop by the learner's own evolving beliefs and actions. In EIUR, standard robust learning can converge stably yet form high-confidence, systematically wrong beliefs.

We propose \textbf{metacognitive regulation} as a practical response: a second, introspective control loop that infers experience credibility from \textit{endogenous} evidence in the learner's internal dynamics. We formalize this as a modular \textbf{Monitor--Trust--Regulator (MTR)} decomposition and instantiate it with \textbf{self-diagnosis}, which maintains a slowly varying \textit{experience-trust} variable that softly modulates learning updates, without exogenous reliability labels or an explicit corruption model.

Empirically, in the EIUR regimes studied here, self-diagnosis is associated with improved epistemic identifiability. In reinforcement learning, it enables calibrated skepticism and recovery under systematically corrupted rewards. In supervised learning, it exposes a critical dissociation: \textbf{performance recovery does not imply epistemic recovery}---accuracy can rebound while internal belief dynamics remain locked-in by early misleading data, a failure detectable only through introspective diagnostics. Together, MTR and self-diagnosis provide an organizing abstraction and a concrete design template for intrinsic reliability assessment in autonomous learning under unobservable reliability.
\end{abstract}

\keywords{metacognitive regulation; epistemic identifiability; self-diagnosis; introspective learning; experience trust}

%%%%%%%%%%%%%%%%%%%%%%%%%%%%%%%%%%%%%%%%%%%%%%%%%%%%
% Main content
% Option A (recommended): keep your whole paper in content.tex
%%%%%%%%%%%%%%%%%%%%%%%%%%%%%%%%%%%%%%%%%%%%%%%%%%%%
\section{Introduction}

The hallmark of an autonomous intelligent agent is its ability to form reliable beliefs from experience. Contemporary machine learning largely frames this as an \emph{optimization} problem: update internal parameters to improve performance under observed feedback. This framing, however, typically assumes that the incoming experience is \emph{sufficiently trustworthy} to support belief formation. In controlled benchmarks that assumption often holds; in open-world settings, feedback can be corrupted, ambiguous, or \textbf{systematically misleading}, and the assumption \textbf{can fail}.

This work focuses on a distinct epistemic regime where that failure is structural rather than incidental. We study \textbf{Epistemic Identifiability under Unobservable Reliability (EIUR)}: (i) each experience carries a latent reliability/credibility that is \emph{not directly observable}; (ii) reliable and unreliable experiences can be \emph{locally indistinguishable} at the level of instantaneous signals; and (iii) data are generated in a \emph{closed loop}, where the agent's current beliefs shape its actions, which shape future feedback. In EIUR, misleading feedback can be incorporated indiscriminately, shaping future data collection and inducing persistent, self-reinforcing error---a form of epistemic captivity.

Most existing approaches address uncertainty through the lens of \textbf{optimization robustness}. Techniques such as gradient clipping \cite{madry2017towards}, entropy regularization, and adversarial training stabilize learning under noise and outliers. Related approaches include: \textit{robust optimization} methods that bound gradient norms or modify loss surfaces \cite{zhang2019theoretically}; \textit{uncertainty quantification} techniques that estimate predictive confidence \cite{gal2016dropout,kendall2017uncertainties,lakshminarayanan2017simple}; \textit{learning with noisy labels} methods that identify and correct mislabeled examples \cite{han2018co}; and \textit{meta-learning} approaches that adapt learning strategies based on experience \cite{finn2017model,andrychowicz2016learning}. Yet these methods often share an implicit premise: that \emph{all} incoming experience, however noisy, should contribute to learning. They answer ``\textit{how} to learn stably'' but do not decide ``\textit{whether} to learn from this experience at all.'' Consequently, a system can be stable and convergent, yet converge with high confidence to a systematically wrong model of the world. This exposes a gap that is not merely about robustness, but about \textbf{epistemic identifiability}: whether the learner can reliably distinguish informative from systematically misleading experience when reliability is unobservable.

We argue that closing this gap requires equipping learning systems with a second, introspective capability: \textbf{metacognitive regulation}. Concretely, metacognitive regulation is an internal control loop that (i) monitors the learning process, (ii) infers the credibility of the driving experiences, and (iii) regulates their influence on parameter updates. In cognitive science, such ``thinking about thinking'' is central to judgment and decision-making \cite{fleming2017metacognitive}; here we study it as a computational design principle for EIUR-like learning systems.

Inspired by introspective regulation in biological systems, we propose a computable, modular framework---the \textbf{Monitor--Trust--Regulator (MTR) loop}---for intrinsic reliability assessment under unobservable feedback credibility. MTR does \emph{not} rely on exogenous reliability labels or an explicit parametric corruption model. Instead, it leverages \emph{endogenous} signals generated by the learner itself, under the assumption that internal learning dynamics contain slowly varying information about experience credibility---an assumption we study empirically in the EIUR regimes considered here. The MTR decomposition separates three roles: \textbf{Monitor} (extract internal descriptors), \textbf{Trust} (aggregate descriptors into an experience-credibility estimate), and \textbf{Regulator} (softly modulate learning gain to control how strongly experience updates beliefs).

We instantiate MTR with a lightweight mechanism we call \textbf{self-diagnosis}. In reinforcement learning, self-diagnosis monitors internal learning descriptors (e.g., policy drift and uncertainty) to maintain a slowly varying \emph{experience-trust} variable that softly modulates updates, enabling calibrated skepticism and recovery under systematically misleading reward signals. In supervised learning, self-diagnosis uses introspective diagnostics (e.g., predictive entropy dynamics) to reveal epistemic failures that remain invisible under performance-only evaluation: notably, \textbf{performance recovery does not imply epistemic recovery}. Across both paradigms, we provide empirical evidence that self-diagnosis is critical for epistemically identifiable learning in the EIUR regimes considered.

\paragraph*{Contributions.}
The contributions of this work are fourfold:
\begin{enumerate}[label=(\roman*)]
    \item \textbf{Problem formulation.} We formalize \textbf{EIUR} as a learning regime characterized by latent (unobservable) experience reliability, local indistinguishability, and closed-loop data generation, and clarify why this creates an identifiability gap beyond conventional robustness.
    \item \textbf{Framework.} We introduce the \textbf{Monitor--Trust--Regulator (MTR) loop} as a \textbf{modular architectural decomposition} for intrinsic reliability assessment from endogenous learning dynamics.
    \item \textbf{Instantiation.} We instantiate MTR as a lightweight, interpretable \textbf{self-diagnosis} mechanism that maintains a slowly varying \emph{experience-trust} signal and uses it to softly regulate learning updates.
    \item \textbf{Empirical evidence across paradigms.} 
    We provide evidence that self-diagnosis is empirically associated with improved epistemic identifiability in the EIUR regimes examined here:
    \begin{itemize}
        \item In \textbf{reinforcement learning}, we demonstrate the full MTR regulatory loop and show that self-diagnosis enables calibrated skepticism and recovery under systematically corrupted rewards.
        \item In \textbf{supervised learning}, we validate the \textbf{monitoring and diagnostic core} of MTR and identify a dissociation between \emph{performance recovery} and \emph{epistemic recovery}, which we term \emph{epistemic lock-in}.
    \end{itemize}
\end{enumerate}

\paragraph*{Scope and non-claims.}
Our claims are restricted to the EIUR regimes studied in this paper. We do not claim optimality of the particular self-diagnosis instantiation, nor do we claim that endogenous coherence signals suffice in fully adaptive adversarial settings where internal dynamics may carry no usable information about credibility---a challenge shared by many robustness methods \cite{madry2017towards,zhang2019theoretically}. While we demonstrate the full regulatory loop in reinforcement learning, the supervised-learning results are presented as \textbf{diagnostic validation}; closing the loop with an explicit Regulator in supervised learning is left as future work.

Overall, our findings reposition learning under unobservable reliability from an optimization-centric to an epistemology-centric problem. We argue that metacognitive regulation can serve as a practical, endogenous mechanism for intrinsic reliability assessment in EIUR-like settings, and that the MTR decomposition provides an organizing abstraction---and self-diagnosis a concrete template---for building systems that can detect and mitigate certain forms of systematically misleading feedback using endogenous evidence.

\section{A Theoretical Framework for Metacognitive Regulation}

The empirical failures documented in the following sections stem from a common structural cause: the lack of an internal mechanism for epistemic regulation. Here, we formalize this missing capability. We first define the problem class precisely, then introduce a modular architectural decomposition—the Monitor–Trust–Regulator (MTR) loop—that mitigates key EIUR failure modes and improves epistemic identifiability in the regimes studied, and finally distill the design principles that we find useful for mitigating EIUR failure modes in the regimes studied.

\subsection{Problem Formulation: Epistemic Identifiability under Unobservable Reliability}
\label{sec:eiur}

Consider a learning agent interacting with an environment in a closed loop. At time $t$, it takes an action based on its current policy (or model), receives an experience $u_t$ (e.g., a state transition or a labeled datum), and updates its internal parameters $\theta_t$. The experience carries a latent \textit{reliability} $\rho_t \in \{0,1\}$, where $\rho_t=1$ indicates the experience is a faithful reflection of the world's dynamics (or true labels), and $\rho_t=0$ indicates it is systematically misleading. The agent’s goal is to form correct beliefs about the world (or task).

The central difficulty is that $\rho_t$ is \textbf{unobservable}. The agent sees only the experience $u_t$, not its truthfulness. Furthermore, experiences are \textbf{locally indistinguishable}: a single unreliable experience may be statistically identical to a rare but truthful one. Finally, learning is \textbf{sequential and belief-dependent}: current beliefs influence future actions, which in turn influence the distribution of future experiences.

We define the problem of \textbf{Epistemic Identifiability under Unobservable Reliability (EIUR)} as follows: can the agent’s sequence of beliefs $\{\theta_t\}$ converge to the correct model of the world, given that it never observes $\rho_t$ and that the data-generating process depends on its own actions?

\begin{proposition}[ Epistemic Identifiability Gap under EIUR]\label{prop:gap}
Consider the EIUR setting defined above, where (i) experience reliability $\rho_t$ is latent and unobservable, (ii) individual reliable/unreliable experiences are locally indistinguishable in distribution, and (iii) the data-generating process depends on the agent’s own actions (closed-loop distributional shift).
Under these conditions, learning algorithms that rely \emph{solely} on instantaneous performance signals (e.g., reward, loss, gradient norms) \textbf{may fail to identify the underlying reliability in certain problem instances}. Specifically, there exist EIUR instances in which such an algorithm converges stably—and may even achieve high performance—while forming systematically biased beliefs that are epistemically unrecoverable without access to exogenous reliability labels.
This gap highlights the potential value of a metacognitive regulatory mechanism that leverages endogenous, slow-timescale evidence about the coherence of learning dynamics, rather than instantaneous error signals alone.
\end{proposition}

\textbf{Intuition}: Instantaneous signals cannot distinguish between “the world is inherently stochastic” (high variance, unbiased) and “the feedback is systematically lying” (low variance, biased). Robustness techniques suppress large updates but do not question the epistemic status of the data itself. The agent needs a source of evidence that is orthogonal to the immediate feedback—a source we argue must be \textit{endogenous} to the learning process.

\subsection{The Monitor–Trust–Regulator (MTR) Framework}

\begin{figure}[t]
    \centering
    \includegraphics[width=0.95\linewidth]{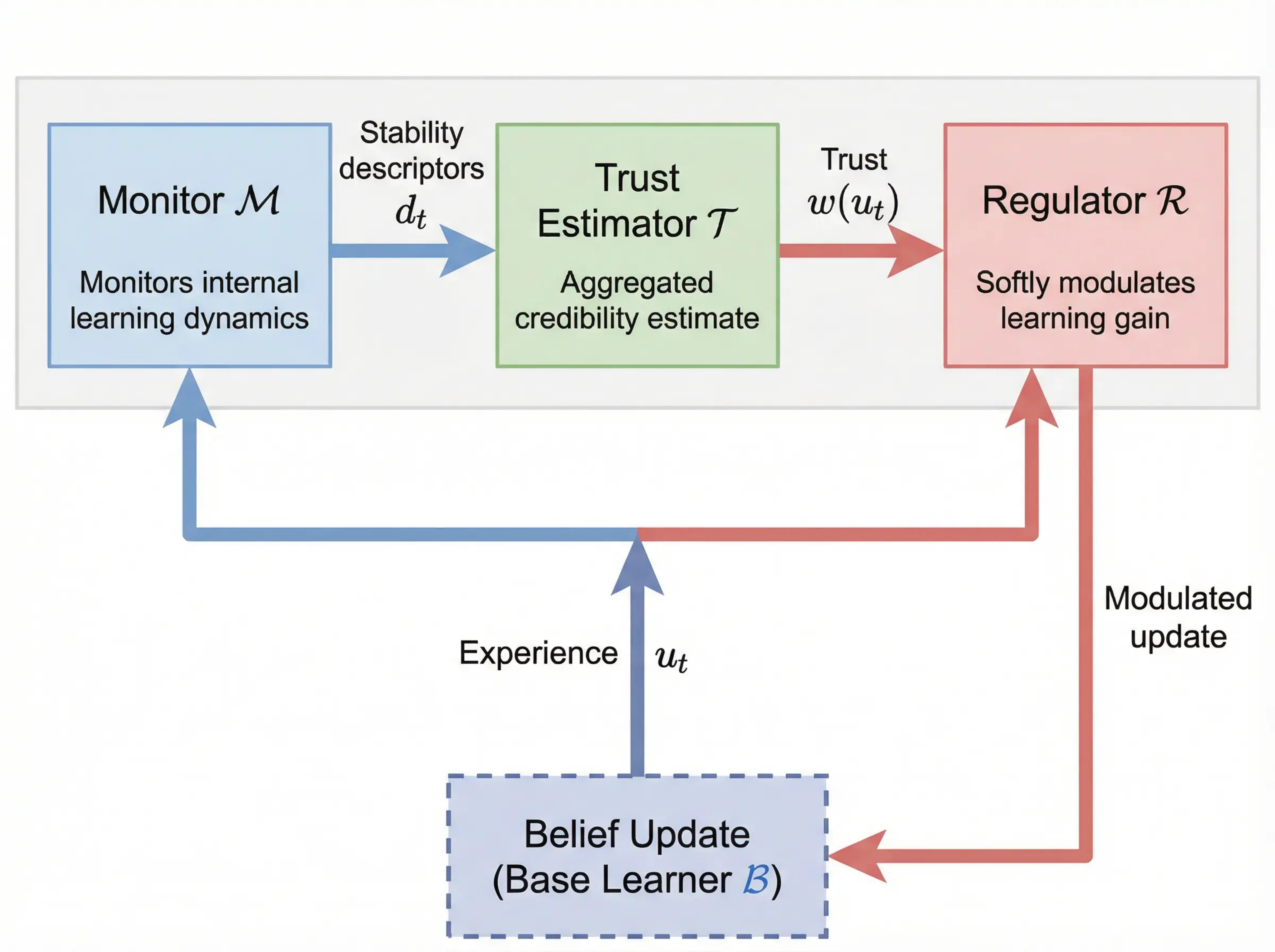}
    \caption{\textbf{The Monitor--Trust--Regulator (MTR) framework for metacognitive regulation.} 
    A separable regulatory loop (top) operates alongside the primary learning loop (bottom). 
    The Monitor $\mathcal{M}$ extracts stability descriptors from internal learning dynamics. 
    The Trust Estimator $\mathcal{T}$ aggregates these descriptors into a slow-varying assessment of experience credibility. 
    The Regulator $\mathcal{R}$ uses this assessment to softly modulate the influence of experience on the base learner $\mathcal{B}$. 
    This introspective loop mitigates epistemic lock-in and improves identifiability in the EIUR regimes studied, 
    without requiring exogenous reliability labels.}
    \label{fig:mtr_framework}
\end{figure}

To improve epistemic identifiability, the agent requires a separable regulatory loop that operates on a different objective: evaluating the credibility of the learning process itself. We propose the MTR framework, illustrated in \autoref{fig:mtr_framework}.

% ============================================================
% Fig. 1: Monitor--Trust--Regulator (MTR) framework
% ============================================================

The MTR framework is a \textbf{separable, modular regulatory loop} that can be embedded into many existing learning algorithms. It consists of three core components that form a closed loop:
\begin{enumerate}
    \item \textbf{Monitor ($\mathcal{M}$)}: An internal sensor that extracts \textit{temporal descriptors} $\mathbf{d}_t$ from the evolution of the agent's internal states over a recent window of experience. These descriptors quantify properties like belief consistency, policy stability, or uncertainty fluctuations---signals that reflect how coherently the agent is learning from a given experience stream.
    
    \item \textbf{Trust Estimator ($\mathcal{T}$)}: A slow integration process that maps the stream of descriptors $\{\mathbf{d}_t\}$ to a \textit{trust weight} $w(u) \in [0,1]$ for each experience (or experience type). Crucially, $\mathcal{T}$ operates on a \textit{slower timescale} than the base learner's parameter updates. This timescale separation allows it to form a stable, aggregated judgment about experience reliability, filtering out transient noise.
    
    \item \textbf{Regulator ($\mathcal{R}$)}: A gain controller that applies the trust weight $w(u)$ to modulate the contribution of experience $u$ to the base learner's update. The modulation is \textit{soft} (preserving gradient information) and \textit{reversible} (allowing trust to be revised with new evidence).
\end{enumerate}

\textbf{Contrast with existing approaches.} Unlike adversarial training which modifies the loss surface \cite{madry2017towards} or ensemble methods which average multiple models \cite{lakshminarayanan2017simple}, MTR regulates learning gain without altering the base objective or architecture. Unlike confidence calibration techniques that post-process predictions \cite{guo2017calibration}, MTR operates online during training. This soft, online modulation shares philosophical similarities with meta-learning's adaptation of learning rates \cite{finn2017model} but focuses specifically on experience credibility rather than task adaptation.

Formally, if the base learner $\mathcal{B}$ has an update rule $\theta_{t+1} = \theta_t - \eta \nabla L(u_t; \theta_t)$, the MTR-augmented update becomes:
\[
\theta_{t+1} = \theta_t - \eta \; w(u_t) \; \nabla L(u_t; \theta_t),
\]
where $w(u_t) = \mathcal{T}(\mathcal{M}(\{u_{t-W}, \dots, u_t\}; \theta))$. The MTR loop is \textbf{abstraction-level agnostic; we instantiate it here for gradient-based learners} without modifying their internal architecture or objective. The framework is designed to address the EIUR problem, with its core design principles being the use of \emph{endogenous evidence}, \emph{timescale separation}, and \emph{calibrated reversibility}.

\subsection{Design Principles Derived from EIUR Structure}

The structural constraints of EIUR imply three essential design requirements for any metacognitive regulation mechanism:

\textbf{(1) Endogenous Evidence.} Diagnostic signals must originate from internal learning dynamics rather than exogenous supervision, as reliability is unobservable. External labels violate the premise of autonomy and cannot address epistemic identifiability.

\textbf{(2) Timescale Separation.} 
The trust variable should evolve on a slower timescale than the base learner’s parameter updates, so that credibility inference is based on temporally aggregated evidence rather than instantaneous fluctuations. 
Without sufficient separation, trust can become overly reactive and noisy, coupling to short-term learning variance and destabilizing regulation; with overly slow updates, it may fail to adapt to regime shifts.

\textbf{(3) Calibrated Reversibility.} Regulation must act as a soft, revisable gain rather than an irreversible filter, enabling adaptation under non-stationary reliability. Hard filtering risks irreversible information loss from early misjudgments.

A fourth emergent requirement is \textbf{Non-Confusion}: the system must distinguish epistemic difficulty (noisy but unbiased feedback) from epistemic unreliability (systematic bias). These principles motivate the MTR design and are validated empirically in subsequent sections.

\subsection{Cross-Paradigm Instantiation: From Framework to Mechanism}

The MTR framework is a template rather than a single algorithm. Instantiating it requires selecting concrete implementations for the Monitor $\mathcal{M}$, Trust Estimator $\mathcal{T}$, and Regulator $\mathcal{R}$ that match the learning paradigm and the available endogenous signals.

\begin{table}[t]
\centering
\caption{\textbf{Cross-paradigm view of the MTR framework.} The \emph{architecture} (Monitor--Trust--Regulator) is invariant; only the monitored signals and update interfaces depend on the learning paradigm. In this work, we demonstrate the full regulatory loop in RL, and validate the monitoring/diagnostic core in SL.}
\label{tab:mtr_instantiation}

\begin{tabularx}{\linewidth}{p{0.20\linewidth} X X}
\toprule
\textbf{Component} 
& \textbf{Reinforcement Learning (Policy Gradient)} 
& \textbf{Supervised Learning (Classification)} \\
\midrule

\textbf{Monitor ($\mathcal{M}$)} 
& Policy drift $\Delta \pi(s)$; action consensus; entropy variance $\mathrm{Var}(H(\pi))$
& Predictive entropy trajectory $H(p(y|x))$ \emph{(implemented; diagnostic signal)} \\

\textbf{Trust Estimator ($\mathcal{T}$)} 
& Unsupervised clustering of descriptors (e.g., GMM) over a sliding window; trust given by posterior probability of a ``stable'' cluster
& Temporal smoothing of entropy into a phase-level risk/trust score \emph{(implemented; diagnostic)} \\

\textbf{Regulator ($\mathcal{R}$)} 
& Trust-weighted policy-gradient update  
$\mathbb{E}\!\left[w(u)\,\nabla_\theta \log \pi(a|s)\,A(s,a)\right]$
& Trust-weighted cross-entropy  
$\mathbb{E}\!\left[w(u)\,\mathcal{L}_{\mathrm{CE}}(y,\hat{y})\right]$  
\emph{(conceptual extension; not activated in this work)} \\

\textbf{Core signal} 
& \multicolumn{2}{X}{\textit{Belief-dynamics coherence}: temporal consistency of internal learning dynamics in response to experience.} \\

\bottomrule
\end{tabularx}
\end{table}

As summarized in \autoref{tab:mtr_instantiation}, the same abstract decomposition guides instantiation across reinforcement and supervised learning. In RL, the monitor extracts descriptors from policy/value learning dynamics and the regulator directly modulates the policy update. In SL, we focus on the monitoring and trust-estimation components to validate that endogenous belief-dynamics signals can render epistemic failures observable even when performance metrics recover. In both paradigms, the unifying epistemic signal is the \textit{coherence of belief dynamics} over time, rather than any single instantaneous reward or loss value.

In the following sections, we instantiate MTR through a specific, interpretable mechanism we call \textbf{self-diagnosis}. Our goal is not to claim optimality of this instantiation, but to demonstrate that mechanisms adhering to the MTR principles can, in practice, mitigate key EIUR failure modes and improve epistemic identifiability in the regimes studied. Empirically, we show that the absence of such regulation is associated with epistemic unidentifiability, whereas its presence promotes calibrated skepticism, reversibility, and stable belief formation in our testbeds.

\section{Empirical Validation: The Necessity of Metacognitive Regulation}

The theoretical framework above posits that metacognitive regulation is necessary for epistemically identifiable learning under unobservable reliability. We now present empirical evidence that validates this claim. We instantiate the MTR framework via a concrete mechanism—\textbf{self-diagnosis}—and test its core predictions. Rather than focusing on final performance, these experiments probe the \textit{epistemic mechanisms} themselves: can beliefs be formed identifiably without self-diagnosis? Does the system develop a stable trust variable? Does trust exhibit the hypothesized properties of calibration and reversibility?

\subsection{Finding 1: Belief formation is epistemically unidentifiable without self-diagnosis}

\begin{figure}[t]
  \centering
  \includegraphics[width=0.8\linewidth]{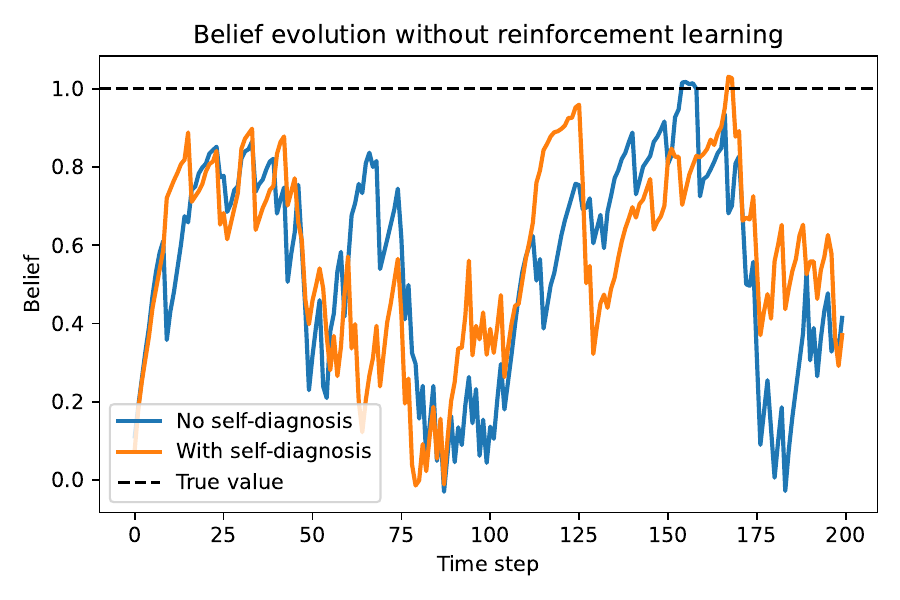}
  \caption{\textbf{Belief evolution under unreliable experience without reinforcement learning.} When experience reliability is unobservable and observations are locally indistinguishable, belief formation is not identifiable without access to temporal structure in learning dynamics. Indiscriminate belief updating leads to persistent instability (blue), whereas self-diagnosis based on belief dynamics enables stable convergence toward the true latent value (orange).}
  \label{fig:belief_dynamics}
\end{figure}

In settings where experience reliability is unobservable and individual observations are locally indistinguishable, we first examine the dynamics of belief formation with and without self-diagnosis. As shown in Fig.~\ref{fig:belief_dynamics}, the contrast is striking: without self-diagnosis, belief trajectories exhibit persistent oscillations and repeated deviations from the true latent value, failing to converge even with repeated exposure to experience. This is not merely transient instability but a structural failure of identifiability—the system cannot determine whether the world is inherently uncertain or whether its experiences are systematically misleading.

With self-diagnosis, belief trajectories gradually stabilize, fluctuations diminish over time, and the system converges reliably toward the true latent value. This comparison demonstrates that self-diagnosis is not merely an optimization technique to accelerate learning, but a \emph{necessary condition for epistemically identifiable belief formation in the EIUR regimes studied}. When experience credibility cannot be directly observed, mechanisms that interpret temporal learning dynamics become essential for distinguishing informative evidence from misleading feedback.

\paragraph*{Claim 1 (Scoped implication).}
Under the EIUR assumptions defined in Section~\ref{sec:eiur}, instantaneous performance signals alone can be insufficient to guarantee epistemically identifiable belief formation. In our EIUR testbeds, learners that update indiscriminately from such signals may converge stably yet remain systematically misled, motivating an additional endogenous, slow-timescale regulatory signal as provided by self-diagnosis.

% \subsection{Finding 2: Self-diagnosis introduces a stable, slowly evolving experience-trust variable}

% To understand how self-diagnosis enables stable belief formation, we examine the internal variable it maintains: \emph{experience trust}. This variable is not a direct function of environmental states or instantaneous rewards; instead, it is inferred from the temporal consistency of learning dynamics. Across all experimental conditions, we observe that experience trust acts as a \emph{slow variable}—remaining relatively stable within training phases and adjusting only gradually when accumulated statistical evidence warrants revision.

% This slow-variable property indicates that experience trust functions as an epistemic regulator rather than a noise detector. It does not react to transient fluctuations but integrates information over time to form calibrated judgments about experience reliability.

\subsection{Finding 2: Self-diagnosis induces a stable, slowly varying experience-trust signal}

To understand how self-diagnosis supports stable belief formation, we examine the internal variable it maintains: \emph{experience trust}. This signal is not a direct function of environmental states or instantaneous rewards; instead, it is inferred from the temporal coherence of endogenous learning dynamics. In the EIUR testbeds studied here, we consistently observe that experience trust behaves as a \emph{slow} (i.e., temporally smoothed) variable relative to the base learner’s parameter updates: it remains approximately stable within training phases and adjusts gradually when sufficient evidence accumulates to warrant revision.

This slow-timescale behavior is consistent with the intended role of experience trust as an epistemic regulator rather than a reactive noise detector. By integrating evidence over time, the trust signal filters short-term fluctuations and yields calibrated, revisable judgments about experience credibility, enabling regulation that is both stable and responsive to sustained changes in reliability.

\subsubsection{Finding 2a: Calibrated skepticism under sustained systematic misleading}
\begin{figure}[t]
  \centering
  \includegraphics[width=0.9\linewidth]{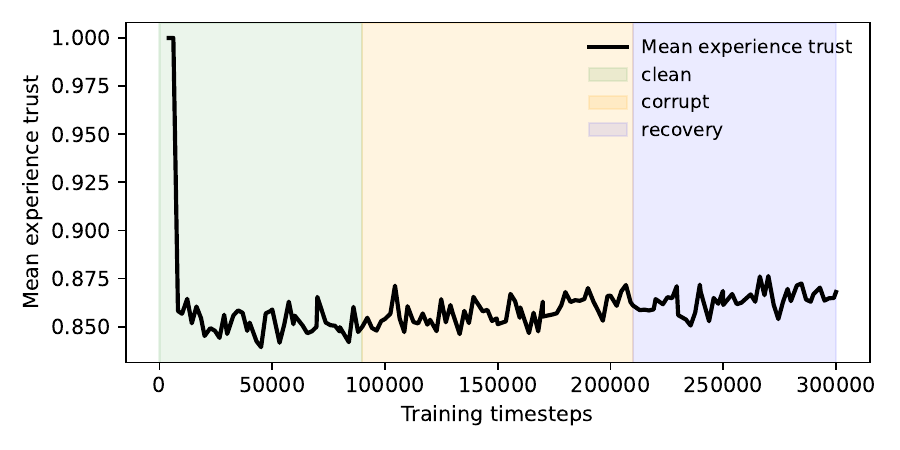}
  \caption{\textbf{Mean experience trust under sustained reward corruption.} Trust decreases during the corruption phase but stabilizes at a suppressed level rather than collapsing. When reliable feedback is restored, trust recovers gradually, demonstrating calibrated skepticism rather than irreversible filtering.}
  \label{fig:trust_corruption}
\end{figure}

Under sustained reward corruption (Fig.~\ref{fig:trust_corruption}), experience trust does not decline indefinitely nor collapse to zero. Instead, it stabilizes at a suppressed yet steady level, indicating that the system maintains a baseline of epistemic skepticism while avoiding catastrophic distrust. When reliable feedback is restored later, trust recovers systematically without requiring external reset signals.

This behavior reveals that experience trust embodies \emph{calibrated skepticism}: it reduces confidence in systematically misleading experiences while preserving the capacity to revise judgments when evidence changes. This distinguishes it from binary filtering mechanisms that would permanently discard experiences deemed unreliable.

\subsubsection{Finding 2b: Calibrated acceptance under unbiased but ambiguous noise}
\begin{figure}[t]
  \centering
  \includegraphics[width=0.9\linewidth]{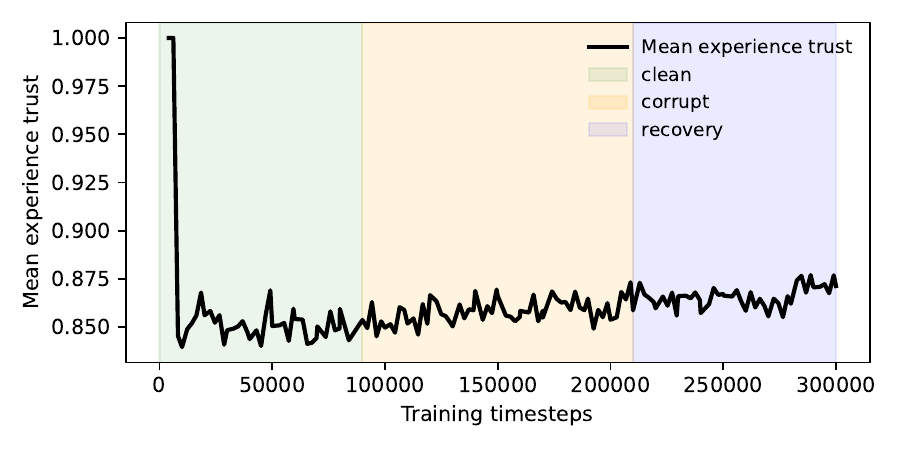}
  \caption{\textbf{Mean experience trust under ambiguous but unbiased reward noise.} Trust forms gradually and stabilizes at moderate levels despite high-variance feedback, indicating that the system distinguishes informative but uncertain experience from systematically misleading experience.}
  \label{fig:trust_ambiguous}
\end{figure}

When feedback contains unbiased noise that increases uncertainty without introducing systematic bias (Fig.~\ref{fig:trust_ambiguous}), experience trust follows a markedly different trajectory. Rather than being suppressed, it forms steadily and stabilizes at moderate-to-high levels. This demonstrates that self-diagnosis does not conflate \emph{epistemic difficulty} with \emph{epistemic unreliability}.

The system recognizes that noisy but unbiased experiences still support coherent belief formation over time and accordingly maintains trust in them. This capacity for \emph{calibrated acceptance} is crucial for learning in realistic environments where perfect signal-to-noise ratios are unavailable.

\subsubsection{Finding 2c: No permanent locking from early misleading experience}
\begin{figure}[t]
  \centering
  \includegraphics[width=0.9\linewidth]{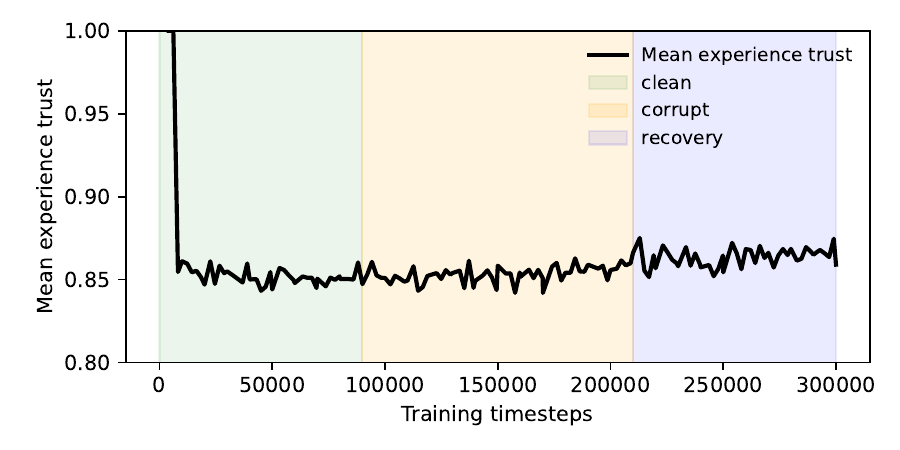}
  \caption{\textbf{Mean experience trust under early misleading rewards.} Trust is initially reduced due to systematically incorrect feedback, but consistently recovers once reliable rewards are restored, indicating that early epistemic errors are revisable rather than permanently fixed.}
  \label{fig:trust_early_misleading}
\end{figure}

A critical test of any epistemic mechanism is whether early errors lead to irreversible consequences. As shown in Fig.~\ref{fig:trust_early_misleading}, when the system encounters systematically misleading feedback early in training, experience trust drops appropriately. However, when subsequent reliable experience becomes available, trust recovers steadily rather than remaining locked at low levels.

This reversibility demonstrates that experience trust is a \emph{dynamic, revisable epistemic quantity}, not a once-formed judgment that permanently biases future learning. Early epistemic errors do not cause irreversible damage to the system's capacity to evaluate experience reliability.

\subsection{Finding 3: Experience-trust stability provides a mechanistic pathway to belief stability}

The stability observed in Fig.~\ref{fig:belief_dynamics} (with self-diagnosis) is not spontaneous; rather, it is \textbf{consistent with} the experience-trust dynamics documented in Findings 2a--2c. Experience trust acts as a regulatory buffer: by attenuating the influence of unreliable experiences while preserving the influence of informative ones, it reduces conflicting update pressures and helps prevent beliefs from being repeatedly pulled in opposing directions.

This pattern \textbf{supports a mechanistic interpretation} in which stable trust modulation underlies stable belief dynamics:
\[
\begin{aligned}
\text{Self-diagnosis}
&\;\rightarrow\; \text{Stable experience trust} \\
&\;\rightarrow\; \text{Modulated belief updates} \\
&\;\rightarrow\; \text{Improved epistemic identifiability}
\end{aligned}
\]

The slow-timescale nature of experience trust (Finding 2) suggests that regulation is driven by temporally aggregated evidence rather than transient fluctuations. This enables the system to remain cautious under short-term noise while still adapting when sustained evidence indicates a change in reliability.

\subsection*{Summary of findings}
Together, Findings 1--3 indicate that in EIUR-like settings where experience reliability is unobservable, stable belief formation benefits from an internal epistemic mechanism that evaluates and regulates experience credibility using long-horizon consistency in endogenous learning dynamics. In our testbeds, self-diagnosis implements this mechanism through calibrated skepticism, calibrated acceptance, and revisable (soft) judgments, enabling reliable learning without privileged reliability labels. These empirical results are consistent with the design principles highlighted by the MTR framework: endogenous evidence, timescale separation, and calibrated reversibility.

\section{An Operational Instantiation: The Self-Diagnosis Mechanism}

The findings above validate the necessity of metacognitive regulation. We now present a concrete, lightweight instantiation of the MTR framework—the \textbf{self-diagnosis} mechanism—that can be integrated into standard learning pipelines. This section details its design, demonstrating how the abstract components of Monitor, Trust Estimator, and Regulator are realized in practice, first for reinforcement learning.

\subsection{Algorithmic Overview}

We formalize self-diagnosis as the maintenance of an experience-trust variable that evolves on a slower timescale than policy optimization and regulates the contribution of experience to learning updates. The mechanism operates alongside a base RL algorithm and does not alter its objective, optimizer, or policy architecture.

At a high level, the self-diagnosis mechanism follows a three-stage process: (1) passive monitoring of temporal learning dynamics, (2) extraction of stability-based descriptors that summarize the coherence of these dynamics over time, and (3) trust-modulated learning, in which experience updates are softly reweighted according to inferred credibility. This design induces a closed introspective loop in which internal learning behavior provides epistemic evidence about experience reliability, enabling the agent to attenuate persistently misleading feedback while preserving informative but uncertain experience.

Algorithm~\ref{alg:self_diagnosis} summarizes the self-diagnosis mechanism as a generic, modular procedure that operates alongside a base reinforcement learning algorithm. The algorithm maintains a slowly evolving experience-trust variable that modulates learning updates based on internally inferred credibility. 
Subsequent sections elaborate each component of Algorithm~\ref{alg:self_diagnosis}:
Section~\ref{sec:monitoring} describes the monitoring of temporal learning dynamics (lines~6–7),
Section~\ref{sec:descriptors} details the construction of stability-based descriptors (line~8),
Section~\ref{sec:trust} specifies how experience trust is estimated and applied (lines~9–10),
and Section~\ref{sec:scope} discusses scope and generality.

% original Alogrithm 1 （存在编译错误）
\begin{algorithm}[t]
\caption{Self-Diagnosis Module (Generic Form)}
\label{alg:self_diagnosis}
\begin{algorithmic}[1]
\Require Base RL algorithm $\mathcal{A}$ with loss $L(u)$, learning rate $\eta$
\Require Sliding window size $W$, update interval $K$
\State Initialize experience trust $w(u) \leftarrow 1$ for all experiences $u$
\For{each training step $t$}
    \State Collect experience $u_t = (s_t, a_t, r_t, s_{t+1})$
    \State Update the base learner using a trust-modulated loss $L_{\text{SD}}$ (Eq.~\ref{eq:trust_loss})
    \If{$t \bmod K = 0$}
        \State Monitor recent learning dynamics over window $W$
        \State Extract stability descriptors (e.g., policy drift, action consensus, entropy variance)
        \State Estimate experience trust $w(u) \in [0,1]$ from descriptor coherence
    \EndIf
\EndFor
\end{algorithmic}
\end{algorithm}

\begin{equation}
\label{eq:trust_loss}
L_{\text{SD}}
= \mathbb{E}_{u \sim \mathcal{D}}\!\left[w(u)\cdot L(u)\right].
\end{equation}
where $L_{\text{SD}}$ denotes the self-diagnosis–modulated learning objective.

\subsection{Experience-Level Monitoring}
\label{sec:monitoring}
In a standard RL setting with transition tuples \(u=(s,a,r,s')\), we instrument the learning process to record policy- and value-related quantities associated with state \(s\) over time. A sliding temporal window captures recent observations of these signals, characterizing short- to medium-term learning dynamics with minimal overhead.

\subsection{Stability-Based Reliability Descriptors}
\label{sec:descriptors}
We summarize monitored dynamics into three interpretable descriptors:
\begin{itemize}
    \item \textbf{Policy Drift}: Changes in policy distribution across updates
    \item \textbf{Action Consensus}: Temporal consistency in preferred action
    \item \textbf{Entropy Variance}: Fluctuations in policy uncertainty over time
\end{itemize}
These descriptors form a reliability signature indicating how the agent's internal beliefs respond to repeated exposure to an experience. They are concrete realizations of the Monitor’s output $\mathbf{d}_t$ in the MTR framework.

\subsection{Trust Estimation and Modulation}
\label{sec:trust}
Using unsupervised clustering on reliability descriptors, we estimate experience trustworthiness \(w(u)\in[0,1]\). This trust estimate modulates learning through soft reweighting:
\[
\mathcal{L}_{\text{trust}}=\mathbb{E}_{u\sim\mathcal{D}}\big[w(u)\cdot\mathcal{L}(u)\big],
\]
where \(\mathcal{L}(u)\) is the standard loss. This preserves gradient information while attenuating misleading feedback, with trust being revisable as learning proceeds. This implements the Trust Estimator and Regulator components.

\subsubsection{On the update timescale of trust.}
A practical question in MTR instantiation is how frequently the Trust Estimator should update the trust signal.
To probe the temporal dynamics of trust estimation, we perform a sensitivity analysis over the trust-update interval $K$ (i.e., recomputing or refitting the trust estimator every $K$ environment steps / optimization steps, while keeping the base learner unchanged).
Across the EIUR testbeds studied here, we observe a consistent trade-off:
when $K$ is too small (overly frequent updates), the inferred trust signal becomes reactive and noisy, coupling to short-term fluctuations in learning dynamics and yielding unstable modulation;
when $K$ is too large (overly infrequent updates), trust adapts too slowly to changes in feedback reliability, making the learner overly conservative and less responsive during regime shifts.
An intermediate $K$ yields the most stable and informative trust modulation, indicating a ``sweet spot'' that reflects the intended \emph{timescale separation} between the primary learning updates and the metacognitive trust dynamics.
These results suggest that temporal aggregation is beneficial for credibility inference in the EIUR regimes studied, and that trust estimation should be slower than (but not decoupled from) the base learning dynamics.

\subsection{Scope and Generality}
\label{sec:scope}
This instantiation is compatible with both on-policy and off-policy RL algorithms and requires no modification to policy architectures or optimizers. While we use specific descriptors and a clustering estimator, these are incidental to the core claim: self-diagnosis can be operationalized using lightweight, interpretable components, fulfilling the promise of the MTR framework.

The proposed self-diagnosis mechanism is agnostic to the learning paradigm. While instantiated here for RL, its core principle — monitoring internal dynamics to infer experience trust — is a \textbf{widely applicable introspective mechanism} applicable to many learning systems that update beliefs from sequential experience. In Section~\ref{sec:sl_diagnostic}, we apply this same capability to a supervised learning setting to diagnose epistemic failure under structured label bias, demonstrating its generality beyond reinforcement learning (we only demonstrate PPO-style policy gradient RL and a diagnostic SL study here).

\section{Reinforcement Learning Validation}

To demonstrate that self-diagnosis is not merely a theoretical construct but functions in practical learning systems, we evaluate its instantiation in standard RL environments under reward corruption. These experiments serve as \emph{validation} that the epistemic mechanisms identified in Section 3 translate to improved learning stability.

% \begin{figure}[t]
%     \centering
%     \includegraphics[width=0.8\linewidth]{figs/learning_curve.pdf}
%     \caption{\textbf{Learning curves on HalfCheetah-v4 under reward corruption} ($p_r = 0.3$). PPO with self-diagnosis (PPO+SD) achieves higher final performance with consistently reduced variance compared to standard PPO, reflecting improved learning stability. Shaded regions denote standard deviation across 5 random seeds.}
%     \label{fig:learning_curve}
% \end{figure}

\subsection{Improved Stability Under Corrupted Feedback}

% As shown in Fig.~\ref{fig:learning_curve}, PPO augmented with self-diagnosis (PPO+SD) exhibits substantially improved learning stability on HalfCheetah-v4 under reward corruption. 

Table~\ref{tab:robustness_p03} quantifies these improvements: PPO+SD achieves higher mean return, lower variance, and significantly better worst-case performance. This validation confirms that the epistemic benefits of self-diagnosis—calibrated trust, reversibility, and stability—translate to practical learning advantages in noisy environments.
The reduced variance and higher final return indicate that the epistemic regulation provided by self-diagnosis translates directly to more robust optimization.

% 优化宽度
\begin{table}[t]
\centering
\caption{Final performance on HalfCheetah-v4 under reward corruption ($p_r = 0.3$). PPO+SD improves mean return, reduces variance, and strengthens worst-case performance. (Table reports mean$\pm$std across seeds at the final evaluation point, whereas Appendix Table~\ref{tab:ablations} reports mean/std computed over the final ten evaluation checkpoints and may use different ablation variants.)}
\label{tab:robustness_p03}

\begin{tabularx}{\linewidth}{
  >{\raggedright\arraybackslash}X
  >{\centering\arraybackslash}X
  >{\centering\arraybackslash}X
}
\toprule
Method & Mean Return & Worst-case \\
\midrule
PPO & $804.6 \pm 136.6$ & 383.1 \\
\textbf{PPO+SD (Ours)} & $\mathbf{854.4 \pm 103.3}$ & $\mathbf{633.5}$ \\
\bottomrule
\end{tabularx}

\end{table}

\subsubsection{Experiment 3 (Structured corruption): State-dependent reward corruption.}
To address the concern that our method may only be effective under uniformly random reward noise,
we evaluate a structured corruption setting where reward corruption is \emph{state-conditional}.
During the corruption phase, reward replacement is triggered only when the current state falls into a predefined region
(e.g., defined by an observable state coordinate exceeding a threshold); \emph{conditioned on being triggered},
the true reward is replaced by a random value with probability $p_r = 0.3$.
% This differs from random corruption, where reward replacement occurs independently of the state.
% We report the mean return and variability (standard deviation) aggregated across seeds following the Table~2 protocol.

We report mean return and variability (standard deviation) across seeds. Results are summarized in Table~\ref{tab:state_dep_corruption}.

\begin{table}[t]
\centering
\caption{State-dependent reward corruption stress test on HalfCheetah-v4.
Numbers follow the Table~\ref{tab:robustness_p03} aggregation protocol (seed-wise last-10 statistics, then aggregated across seeds).}
\label{tab:state_dep_corruption}

\small
\begin{tabularx}{\linewidth}{
  >{\raggedright\arraybackslash}X
  >{\centering\arraybackslash}X
  >{\centering\arraybackslash}X
}
\toprule
Method & Mean Return $\uparrow$ & Std $\downarrow$ \\
\midrule
PPO & 1054.0 & 181.6 \\
SD-PPO & 1914.3 & 158.4 \\
\bottomrule
\end{tabularx}

\end{table}

\subsection{Practicality and Interpretability}

Beyond performance, self-diagnosis offers interpretability through its stability descriptors, which explain why specific experiences are downweighted. Computationally, it introduces minimal overhead (approximately 4.2\% increased training time, see Appendix Table~\ref{tab:ablations}), making it feasible for real-world deployment.

\section{Generalization to Supervised Learning: A Diagnostic Study of Epistemic Lock-in}
\label{sec:sl_diagnostic} 
The MTR framework posits that metacognitive regulation is a \textit{general} capability, not specific to reinforcement learning. To test this, we \textbf{instantiate the \emph{monitoring and diagnostic} components of the MTR framework} in a supervised learning (SL) setting under \textbf{structured label bias}. This diagnostic experiment isolates a critical prediction: that \textbf{performance recovery does not imply epistemic recovery}. A model can learn to output correct labels (high accuracy) while its internal belief formation remains irreversibly distorted—a state we call \textit{epistemic lock-in}. By monitoring internal dynamics such as predictive entropy, self-diagnosis makes this hidden failure \textbf{identifiable}, thereby validating the \emph{diagnostic} power of the MTR framework across learning paradigms.

\subsection{Experimental Setup}

We designed a structured bias experiment where the first 30\% of the training steps were corrupted with systematic label misassignments, and the remaining 70\% used clean labels. This setup simulates a learning scenario in which early experience is consistently misleading but later corrected, allowing us to examine whether epistemic failure can be detected independently of final task performance.

\subsubsection{Training Details}

The model is a two-layer MLP with a hidden layer of 256 units and ReLU activation, followed by a 10-class softmax output. The network is trained for 20,000 steps using the Adam optimizer with a learning rate of $1e\text{-}3$ and a batch size of 128.

Structured bias is introduced deterministically during the first 6,000 steps (30\% of training), where labels `3` are flipped to `8` and `5` to `6`. Clean labels are restored afterward, allowing us to assess whether the model epistemically “recovers” or remains locked into early misleading beliefs despite final performance improvements.

\subsubsection{Self-Diagnosis Logging}

In the self-diagnosis configuration, we compute the predictive entropy of the model's softmax output distribution for each batch:
\[
H(p) = -\sum_{i=1}^{K} p_i \log p_i
\]
where \( p_i \) is the predicted probability for class \( i \). The average entropy per batch is logged at every training step but is not used for backpropagation—it serves purely as an epistemic diagnostic signal (an instantiation of the Monitor).

These entropy values are later used to compute standard diagnostic metrics including AUROC, Brier score, and negative log-likelihood (NLL), treating each training step as either "biased" (early misleading phase) or "clean" (later phase). This evaluation quantifies whether the entropy signal can distinguish epistemic failure over time.

\subsection{Results}

\begin{figure}[h!]
    \centering
    \includegraphics[width=0.7\textwidth]{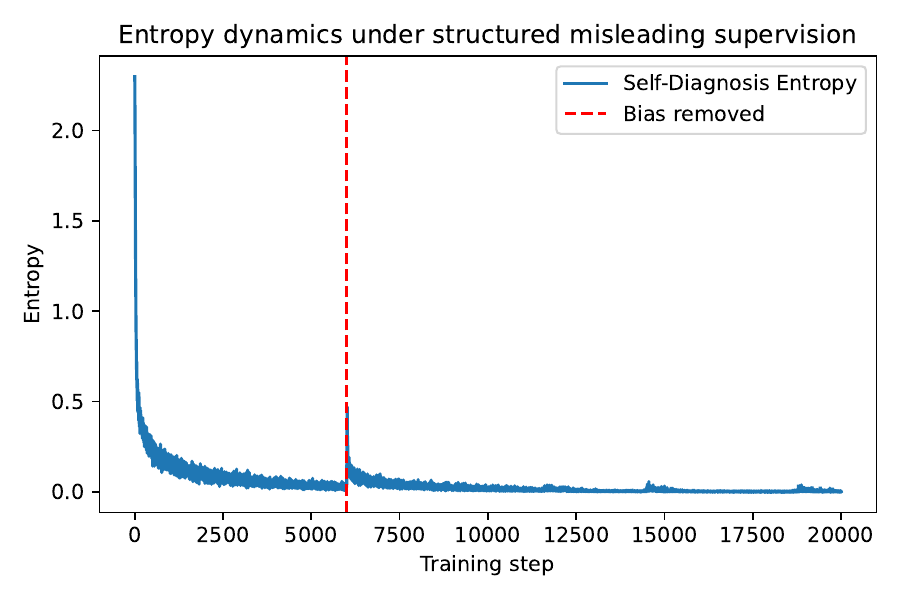}
    \caption{Entropy dynamics under structured misleading supervision. The red vertical line marks the point where the structured bias is removed.}
    \label{fig:fig_structured_entropy}
\end{figure}

Model performance was evaluated using clean test accuracy, while epistemic dynamics were assessed through the evolution of predictive entropy during training. As shown in \autoref{fig:fig_structured_entropy}, entropy decreased rapidly during the misleading phase and remained low even after the structured bias was removed.

Despite the restoration of clean supervision, the entropy trajectory does not exhibit a rebound or reset. This indicates that, although the model ultimately recovers high clean accuracy (approximately 98\%), the epistemic effects of early misleading experience persist. Performance recovery therefore does not imply epistemic recovery.

\subsection{Quantifying Epistemic Identifiability}

While \autoref{fig:fig_structured_entropy} demonstrates that epistemic failure leaves a persistent internal trace, we further assess whether this trace can serve as a usable diagnostic signal. To this end, we evaluate the predictive entropy as a classifier for identifying whether samples originate from the biased training phase.

\begin{figure}[h!]
    \centering
    \includegraphics[width=0.55\textwidth]{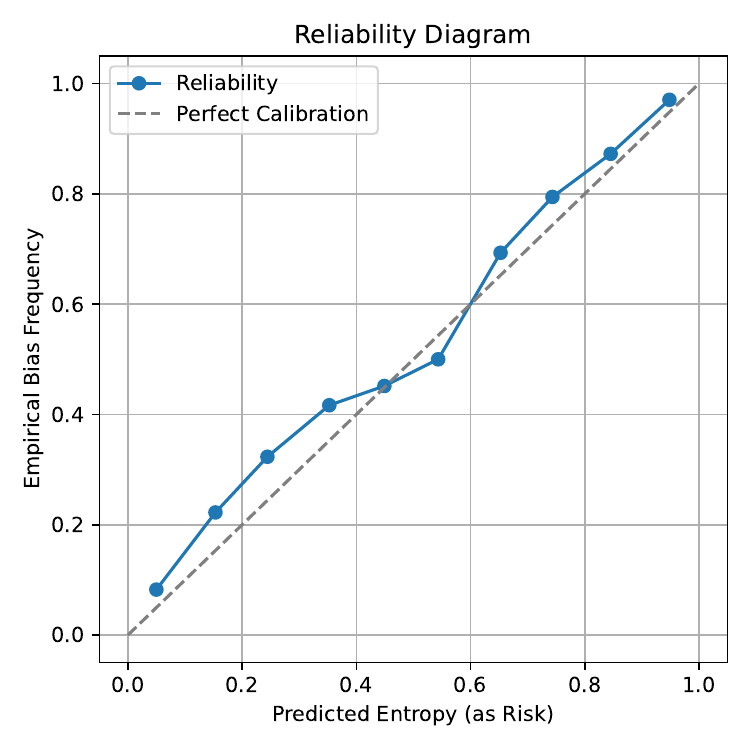}
    \caption{Reliability diagram for entropy-based self-diagnosis under structured bias. Predictive entropy is treated as a risk score indicating epistemic unreliability.}
    \label{fig:fig_entropy_reliability}
\end{figure}

As shown in \autoref{fig:fig_entropy_reliability}, entropy provides a meaningful diagnostic signal: it achieves an AUROC of 0.81 for distinguishing biased from clean training phases, indicating that epistemic failure is not only observable but also quantitatively identifiable. Calibration metrics (Brier score and negative log-likelihood) further show that this diagnostic signal is informative, though not perfectly calibrated, reflecting the fact that entropy is a passive diagnostic rather than an explicitly trained classifier.

\subsection{Discussion: Diagnostic Contribution and Scope}

In this supervised‑learning experiment, we \textbf{instantiate and validate the Monitor (M) and Trust Estimator (T) components} of the MTR framework, demonstrating their ability to \textbf{diagnose epistemic lock‑in} even when performance metrics recover. The \emph{Regulator (R)}, which would actively modulate learning updates based on the inferred trust, is \textbf{not activated here}; its integration for belief correction in SL is a natural extension left for future work. This diagnostic study therefore serves two purposes: (i) it reveals the dissociation between performance and epistemic recovery, a phenomenon that conventional evaluation would miss; and (ii) it provides cross‑paradigm evidence that the \emph{monitoring and trust‑estimation} principles of the MTR framework are applicable beyond RL, offering a general tool for \textbf{internal reliability assessment} in learning systems.

These results reinforce a central claim of this work: \textbf{performance recovery does not imply epistemic recovery}. Even when final accuracy matches that of a cleanly trained model, the internal epistemic state may remain irreversibly shaped by early misleading experience.

Importantly, self-diagnosis does not aim to repair or undo epistemic failure in the supervised setting. Instead, its role is to render epistemic failure \emph{identifiable}. Without self-diagnosis, a model that recovers high performance offers no internal indication that its learning process was ever compromised. With self-diagnosis, the formation and persistence of epistemic lock-in become observable and quantifiable.

Together with the reinforcement learning results presented earlier, this experiment demonstrates that self-diagnosis addresses a fundamentally epistemic problem rather than a purely optimization-based notion of robustness. In supervised learning, epistemic failure may be masked by performance recovery; in reinforcement learning, the same failure directly constrains exploration and degrades performance. In both cases, self-diagnosis provides the internal signals necessary for cognitively transparent learning systems to recognize the reliability of their own learning dynamics. This cross-paradigm evidence supports the MTR framework as an organizing abstraction for metacognitive regulation in the studied regimes.

\section{Discussion: Towards a Science of Metacognitive AI}

Our results motivate a reframing of learning under unobservable feedback reliability: beyond ensuring \emph{optimization robustness}, an autonomous learner must also maintain \emph{epistemic identifiability}---the ability to avoid converging stably to systematically wrong beliefs when experience credibility is latent and data are generated in a closed loop. Within the EIUR regimes studied in this paper, we show that incorporating an introspective control loop can expose and mitigate such failures using \emph{endogenous} evidence from learning dynamics. The proposed Monitor--Trust--Regulator (MTR) decomposition, and its self-diagnosis instantiation, provide a concrete starting point for developing this capability.

\subsection{Epistemic Identifiability as a Distinct Research Axis}

A central takeaway is the separation of \emph{optimization robustness} from \emph{epistemic identifiability}. Robustness mechanisms (e.g., gradient clipping, adversarial training) ask: ``How can we learn stably in the presence of noise?'' They typically presume that all experience should contribute to learning, and focus on stabilizing updates. Epistemic identifiability asks a prior question: ``Should this experience influence belief updating at all?'' It concerns the truth-conduciveness of learning when reliability is unobservable, experience sources can be locally indistinguishable, and closed-loop interaction can amplify early mistakes.

Our evidence suggests that, in EIUR-like interactive settings (including those modeled here), systems without an intrinsic credibility assessment mechanism can be vulnerable to self-confirming error even when training dynamics appear stable. This motivates epistemic identifiability as a complementary axis for studying autonomy-relevant learning failures.

\subsection{Metacognitive Regulation as a Computational Design Principle}

We study \emph{metacognitive regulation} as an implementable design principle: a second control loop that monitors internal learning dynamics, aggregates them into a credibility estimate, and regulates learning gain accordingly. The MTR framework formalizes this capability as a separable, modular loop and highlights three implementation-oriented principles: (i) \emph{endogenous evidence} (use internally generated signals rather than exogenous reliability labels), (ii) \emph{timescale separation} (maintain a slowly varying credibility signal), and (iii) \emph{calibrated reversibility} (soft, non-irreversible modulation rather than hard filtering).

We emphasize that MTR is an organizing abstraction rather than a single algorithm. Self-diagnosis is one concrete instantiation; other choices of monitors, trust estimators, and regulators may be more appropriate in different domains. A key open question is when endogenous coherence signals are informative enough to support reliable credibility inference---and when they are not.

\subsection{Potential Application Domains}

The MTR perspective may be useful in domains where feedback reliability is difficult to guarantee and systematic bias can persist, such as interactive robotics in unstructured environments or decision-support systems trained from heterogeneous data sources. In such settings, introspective diagnostics that track the \emph{health} of belief updating may complement conventional performance-based monitoring. At the same time, the applicability of MTR-style regulation depends on whether a domain provides endogenous signals with sufficient temporal structure to support credibility inference; establishing this empirically is necessary before deployment claims can be made.

\subsection{Bridging to Natural Intelligence: Testable Predictions}

Beyond engineering implications, the framework suggests testable hypotheses for cognitive and neural science by positing a slow, integrative trust variable that regulates learning gain.

\textbf{Testable Prediction 1 (Slow modulation of learning gain).} If biological systems implement regulation analogous to an MTR-like loop, we should observe slow, global modulations of learning rate that correlate with the inferred reliability of information sources, beyond instantaneous prediction errors. Neuromodulatory systems are plausible candidates for implementing such slow-variable control.

\textbf{Testable Prediction 2 (Dissociation between performance and metacognition).} Our supervised-learning study shows that behavioral performance can recover while an internal epistemic signal (e.g., entropy dynamics) remains abnormal. This predicts that, under early misleading information, humans may exhibit behavioral correction while metacognitive confidence remains aberrant---a dissociation measurable via confidence reports or physiological proxies.

\subsection{Implications for AI Safety and Auditing}

Unobservable reliability introduces a failure mode that is easy to miss under standard evaluation: a system may appear competent while its internal belief dynamics are biased by systematically misleading experience. In this sense, metacognitive diagnostics can support \emph{auditability} by surfacing epistemic failures that do not immediately manifest as performance drops. While our experiments consider representative corruption patterns, assessing robustness to targeted, adaptive deception requires dedicated study and remains an important direction.

\subsection{Limitations and Open Frontiers}

Our instantiation via self-diagnosis is a proof-of-concept, and several frontiers remain open:

\begin{itemize}
    \item \textbf{Richer Temporal Models}: Our monitor uses simple sliding-window statistics. More powerful temporal models (e.g., state-space models \cite{hochreiter1997long} or attention mechanisms \cite{vaswani2017attention}) could extract more nuanced signatures of epistemic health.
    
    \item \textbf{Integration with Exploration}: In RL, the trust signal could actively guide \textit{where to explore}---connecting to exploration strategies in bandit problems \cite{auer2002finite,lattimore2020bandit} and intrinsic motivation \cite{osband2016deep}.
    
    \item \textbf{Theoretical Bounds}: A formal information-theoretic analysis of the identifiability gap could build on statistical learning theory \cite{vapnik1998statistical} and distribution shift analysis \cite{ben2010theory}.
    
    \item \textbf{Adversarial and Targeted Deception}: While our experiments include structured corruption, assessing MTR against adaptive adversarial strategies \cite{pinto2017robust} remains important for safety applications.
    
    \item \textbf{Scaling to Large Models}: Applying MTR principles to large language models \cite{brown2020language,dohmatob2024metacognitive} and vision transformers \cite{vaswani2017attention} may require new monitoring strategies suited to their distinctive training dynamics \cite{cohen2021gradient}.
\end{itemize}

\subsection{Concluding Position}

This paper does not claim that self-diagnosis is universally optimal, nor that metacognitive regulation resolves all forms of unobservable unreliability. Rather, our results support a narrower claim: in our EIUR testbeds, epistemic identifiability is not guaranteed by optimization robustness alone, and an introspective regulation loop can provide actionable endogenous evidence to diagnose and, in RL, mitigate certain systematic corruption patterns. We view MTR as a practical organizing abstraction for developing such mechanisms and for studying the conditions under which intrinsic credibility assessment is feasible.

\section{Conclusion}

Learning under unobservable reliability raises an epistemic challenge that is distinct from conventional robustness: a system may learn stably yet form high-confidence, systematically wrong beliefs. We formalized this challenge as \textbf{Epistemic Identifiability under Unobservable Reliability (EIUR)} and argued that addressing it requires more than stabilizing updates---it requires mechanisms that infer \emph{whether} experience should influence belief formation.

To this end, we introduced \textbf{metacognitive regulation} and proposed the \textbf{Monitor--Trust--Regulator (MTR)} loop as a modular architectural decomposition for intrinsic credibility assessment from endogenous learning dynamics. We instantiated MTR with a lightweight \textbf{self-diagnosis} mechanism that maintains a slowly varying experience-trust signal and uses it to softly modulate learning gain.

Empirically, within the EIUR regimes studied in this paper, self-diagnosis is associated with improved epistemic identifiability. In reinforcement learning, we demonstrate the full regulatory loop and show calibrated skepticism and recovery under systematically corrupted rewards. In supervised learning, we validate the monitoring/diagnostic core and reveal \emph{epistemic lock-in}: performance can recover while internal belief dynamics remain distorted by early misleading data.

Our claims are intentionally scoped. We do not claim optimality of the specific instantiation, nor do we claim coverage of fully adaptive adversarial settings where endogenous dynamics may provide insufficient information about credibility. Rather, we aim to establish epistemic identifiability as a useful research axis and to provide an organizing abstraction---and a concrete template---for designing and evaluating intrinsic reliability assessment mechanisms in EIUR-like settings. We hope this work motivates further theoretical and empirical study of metacognitive regulation, including richer monitors, learned regulators, and stronger evaluations under targeted deception and domain shift.

\appendix

\section{Experimental Details}

\subsection{Experimental Settings for Epistemic Trust Analysis}

All experiments in Section~3 were conducted in HalfCheetah-v4 using PPO with self-diagnosis, sharing identical network architectures, optimizers, and hyperparameters.
The policy and value networks consisted of two hidden layers of 64 units each with \texttt{tanh} activations.
The learning rate was set to $3\times10^{-4}$ with a linear decay schedule.
Other PPO hyperparameters followed standard practice~\cite{schulman2017proximal}.

\subsubsection{Non-Stationary Reward Corruption}
Training was divided into three phases: clean (first 30\%), corruption (30\%--70\%), and recovery (last 30\%).
During the corruption phase, rewards were corrupted with probability $p_r = 0.3$ by replacing the true reward with a random value sampled from $\mathrm{Uniform}[-1,1]$.
No explicit signal was provided to the agent regarding phase transitions.

\subsubsection{Ambiguous but Informative Feedback}
Rewards were perturbed by zero-mean Gaussian noise,
\[
r' = r + \epsilon, \quad \epsilon \sim \mathcal{N}(0,\sigma^2), \ \sigma = 0.5 ,
\]
resulting in high-variance but unbiased feedback.
This setting increased learning difficulty without introducing systematic unreliability.

\subsubsection{Early Misleading Bootstrap}
To induce an initial epistemic bias, rewards were systematically inverted during the early stage of training.
Specifically, for the first 30\% of training timesteps, the observed reward was given by $r'=-r$.
The true reward signal was restored thereafter for the remainder of training.

\subsection{Robustness Validation in Reinforcement Learning}

\subsubsection{Experimental Setup}
We evaluated self-diagnosis on three continuous-control environments:
\begin{itemize}
    \item \textbf{HalfCheetah-v4}, our primary testbed;
    \item \textbf{Walker2d-v4}, to assess generalization to balance-intensive locomotion;
    \item \textbf{MetaWorld push-v3}, a robotic manipulation task with denser rewards.
\end{itemize}
All environments are commonly used to evaluate robustness in policy optimization~\cite{henderson2018deep}.

Reward corruption was applied as the canonical noise source: with probability $p_r = 0.3$, the true reward was replaced by a random value sampled from a uniform distribution over the reward range observed during clean training.
This setting represents a moderately noisy yet partially informative feedback regime.

\subsubsection{Implementation Details}
Self-diagnosis was implemented using a sliding window of the most recent $N=50$ updates to compute learning-dynamics descriptors.
A two-component Gaussian mixture model was fitted every $K=1000$ environment steps, and the reliability weight was set to the posterior probability of belonging to the stable cluster.
All agents were trained for 300k environment steps and evaluated using deterministic rollouts every 10k steps.
Results were averaged over 5 random seeds.

% \begin{table}[t]
% \centering
% \caption{Implementation-level ablation studies on HalfCheetah-v4 under reward corruption ($p_r=0.3$).
% Results report mean return, standard deviation, and worst-case performance over the final ten evaluation points, averaged across 5 random seeds. (Table 2 in the main text uses the "PPO+SD (KL + Entropy)" variant reported here.)}
% \label{tab:ablations}
% \begin{tabular}{lccc}
% \toprule
% \textbf{Variant} & \textbf{Mean Return} & \textbf{Std. Dev.} & \textbf{Worst-case} \\
% \midrule
% PPO (baseline) & 804.6 & 136.6 & 383.1 \\
% PPO+SD (policy-only) & 854.4 & 28.6 & 633.5 \\
% \midrule
% PPO+SD (KL-only) & 836.5 & 77.6 & 710.7 \\
% PPO+SD (KL + Entropy) & 854.4 & 103.3 & 633.5 \\
% \bottomrule
% \end{tabular}
% \end{table}

\subsubsection{Ablation Studies}
Table~\ref{tab:ablations} summarizes implementation-level ablations examining both where self-diagnosis is applied and which diagnostic signals are used.
Applying self-diagnosis exclusively to the policy loss yields the most stable performance, whereas jointly applying it to both policy and value losses leads to instability.
Regarding diagnostic signals, KL-based policy drift already provides substantial robustness gains, while incorporating entropy variance offers complementary stabilization.
These results indicate that self-diagnosis is most effective when applied selectively and driven primarily by policy-level consistency signals.

\begin{table}[t]
\centering
\caption{Ablation on diagnostic signals for self-diagnosis on HalfCheetah-v4 under reward corruption ($p_r=0.3$).}
\label{tab:ablations}
\begin{tabular}{lccc}
\toprule
\textbf{Variant} & \textbf{Mean Return} & \textbf{Std. Dev.} & \textbf{Worst-case} \\
\midrule
PPO (baseline) & 804.6 & 136.6 & 383.1 \\
\midrule
PPO+SD (KL-only) & 836.5 & 77.6 & 710.7 \\
PPO+SD (KL + Entropy) & 854.4 & 103.3 & 633.5 \\
\bottomrule
\end{tabular}
\end{table}

\subsubsection{Generalization to Additional Domains}
To assess generalization beyond HalfCheetah, we evaluated self-diagnosis on Walker2d-v4 and MetaWorld push-v3 under identical reward corruption settings.
On Walker2d-v4, PPO with self-diagnosis achieved consistently higher mean and worst-case returns than standard PPO, \textbf{suggesting} improved robustness in a balance-sensitive locomotion task.
On MetaWorld push-v3, PPO with self-diagnosis similarly improved average performance under reward corruption.
Although variance remained non-negligible due to task stochasticity, the qualitative improvement trend matched our main results, \textbf{suggesting that the benefits of self-diagnosis may extend beyond locomotion to manipulation domains, though limited to these domains and corruption settings}.

\subsubsection{Practicality and Interpretability}
Self-diagnosis introduces minimal computational overhead, increasing total training time by approximately 4.2\% in HalfCheetah-v4 experiments.
The additional memory required to store learning-dynamics descriptors was approximately 5\% of the replay buffer size.
Beyond efficiency, self-diagnosis improves interpretability by exposing explicit stability descriptors (e.g., policy drift and entropy variance) that explain why particular experiences are downweighted during learning.

%%%%%%%%%%%%%%%%%%%%%%%%%%%%%%%%%%%%%%%%%%%%%%%%%%%%
% Option B: paste your full body here directly
%%%%%%%%%%%%%%%%%%%%%%%%%%%%%%%%%%%%%%%%%%%%%%%%%%%%

% ====== PASTE YOUR SECTIONS STARTING HERE ======
% \section{Introduction}
% ...

%%%%%%%%%%%%%%%%%%%%%%%%%%%%%%%%%%%%%%%%%%%%%%%%%%%%
% References
%%%%%%%%%%%%%%%%%%%%%%%%%%%%%%%%%%%%%%%%%%%%%%%%%%%%

% If you use BibTeX:
% \bibliographystyle{plain}
% \bibliography{refs}

% If you currently use thebibliography in main.tex, keep it:
% \begin{thebibliography}{99}
% ...
% \end{thebibliography}

\end{document}